# Capture Uncertainties in Deep Neural Networks for Safe Operation of Autonomous Driving Vehicles

Liuhui Ding, Dachuan Li*, *Member, IEEE*, Bowen Liu, Wenxing Lan, Bing Bai, Qi Hao*, *Member, IEEE*
Weipeng Cao, *Member, IEEE* and Ke Pei

*Abstract*—Uncertainties in Deep Neural Network (DNN)-based perception and vehicle's motion pose challenges to the development of safe autonomous driving vehicles. In this paper, we propose a safe motion planning framework featuring the quantification and propagation of DNN-based perception uncertainties and motion uncertainties. Contributions of this work are twofold: (1) A Bayesian Deep Neural network model which detects 3D objects and quantitatively capture the associated aleatoric and epistemic uncertainties of DNNs; (2) An uncertainty-aware motion planning algorithm (PU-RRT) that accounts for uncertainties in object detection and ego-vehicle's motion. The proposed approaches are validated via simulated complex scenarios built in CARLA. Experimental results show that the proposed motion planning scheme can cope with uncertainties of DNN-based perception and vehicle motion, and improve the operational safety of autonomous vehicles while still achieving desirable efficiency.

## I. Introduction

Safety is one of the most critical requirements among the criteria for the design and development of Autonomous Driving Vehicles (AV). Most of state-of-the-art AV software stacks are built following the pipeline of perception-decision making-trajectory planning-control, where individual components support the operation of each functionality [1]. Despite significant progress achieved in AV perception and motion planning techniques, new challenges arise when addressing the safety issues induced by data-driven units(e.g. DNN components), due to the lack of effective mechanisms for the evaluation and propagation of uncertainties when integrating various software components.

Due to the effectiveness in extracting features and making predictions from data-rich input [2], Deep Learning (DL) based techniques have been widely adopted in AV systems, especially for perception applications including semantic segmentation [3] and pedestrian and vehicle detection [4]. However, the inherent errors and uncertainties in the output of DL-based perception unit will propagate downstream to decision-making and planning components, which rely heavily on the output of preceding perception modules, and such uncertainties

L. Ding, D. Li, Q. Hao and B. Bai are with the Department of Computer Science and Engineering, Southern University of Science and Technology, Shenzhen, 518055 China e-mail:{11849325, lidc3, haoq, 11510163}@mail.sustech.edu.cn (*Liuhui Ding and Dachuan Li contributed equally to this work. Corresponding author: Dachuan Li, Qi Hao*)

W. Cao is with College of Computer Science and Software Engineering, Shenzhen University, Shenzhen, China 518060, email: caoweipeng@szu.edu.cn

D. Li and Q. Hao are with Sifakis Research Institute for Trustworthy Autonomous Systems, Shenzhen 518055, China

K. Pei is with Huawei Technologies Co. Ltd., Shenzhen 518129, China, email: peike@huawei.com

ultimately affect the planning results. However, existing DL approaches typically output deterministic perception results without explicitly measuring their uncertainty. In addition, although the motion of ego vehicle is supposed to follow the planned reference trajectory, the actual trajectory following result is highly influenced by various driving conditions such as actuator noise, tire adhesion and road surface, resulting in uncertainty of vehicle motion. Besides, the inherent measurement errors also result in uncertainty of the vehicle's motion. Therefore, it is still an open question how to effectively and explicitly evaluate upstream uncertainties and propagate them to the decision-making and planning phase, so as to guarantee operational safety of AVs in the presence of such uncertainties.

To address these challenges, in this paper we present a coherent framework of uncertainty-aware perception and motion planning for AV systems. The primary contributions of this paper can be summarized as follows:

1) We propose **an uncertainty-aware 3D LiDAR object detector (Bayesian Point-RCNN)** which is an extension of the basic Point-RCNN backbone by following the Bayesian Deep Learning (BDL) principal. The detector is able to classify surrounding objects and predict their 3D bounding boxes, while explicitly quantifying both epistemic (EU) and aleatoric uncertainties (AU).
2) We develop **a chance-constrained safe motion planning approach (perception uncertainty-aware-RRT, PU-RRT)** that accounts for perception uncertainties of DNN and motion uncertainty. The algorithm explicitly incorporates DNN perception and motion uncertainty in the form of spatial chance constraints, and generates probabilistic feasible trajectories that bound the operational risk.
3) We provide **implementation details** and conduct **experimental evaluation** of the proposed algorithms in simulated traffic scenarios. Experimental results show an improvement of operational safety by capturing uncertainties of DNN.

The remainder of this paper is organized as follows: Section II reviews the related work in related domains. Section III presents the problem formulations. Section IV provides a detailed description of the 3D LiDAR object detector and the uncertainty-aware motion planning approach. Section V provides experiment results and analysis. Section VI concludes this paper and outlines future work.

## II. RELATED WORK

### A. Uncertainty Quantification for Deep Neural Networks

There are generally two sources of uncertainties in a DNN, namely *epistemic* uncertainty (EU) and *aleatoric* uncertainty (AU) [5], where the former captures uncertainty in the model parameters and the latter relates to the inherent noise of observation inputs. Approaches for the quantification of those uncertainties can be divided into two major categories: sampling-based and non-sampling-based.

Bayesian Deep Learning (BDL) [6] has emerged as a principled paradigm for modeling uncertainties in DNN. Unlike conventional DNN that uses maximum likelihood regression to estimate deterministic network parameters, the BDL framework models weight parameters as stochastic variables that can be estimated by approximating their posterior probability. Sampling-based techniques have been proposed as approximation methods since direct inference of such posterior distributions are normally intractable. Representative methods include Variation inference (VI) [7], [8], [9], which estimate uncertainty by drawing samples of weights from the approximated distributions generated from multiple inference of a single input. Similarly, dropout techniques have also been adopted to approximate uncertainties through multiple forward-passes[10], [11]. Instead of processing the input using the same network architecture, such methods obtain samples from the posterior distributions of weights by randomly dropping out neurons for each forward pass. In contrast, deep ensemble [12] is a non-Bayesian framework and it passes a single input though ensembles of multiple networks to estimate variance of predictions. In contrast to the above sampling-based approaches, non-sampling-based techniques estimate uncertainties via a single forward pass [13], [14], [15]. Such approaches generally require specific design of loss attenuation mechanism or incorporation of additional Gaussian mixture models to the output, making it difficult to converge for detectors with large output space [16].

Recent works attempt to make extensions to DNNs and quantify uncertainties in 3D object detection applications. Feng et al. [17] utilize Monte Carlo dropout technique to quantify uncertainties of a DNN-based framework for 3D vehicle detection tasks. Their approach is further extended in [18] to quantify heteroscedastic AU in a two-stage detector, including AU of region proposal network and refinement subnetwork. Similarly, Meyer et al. [19] extend the LaserNet detector to quantify the uncertainty in vehicle bounding box estimation. However, most existing uncertainty quantification techniques of DNN are designed for improving robustness and accuracy of predictions, without considering the propagation of perception uncertainties to the downstream components to improve overall system safety.

### B. Decision-making and Planning under Uncertainties

Decision-making and planning under environment and state uncertainties can be typically formulated as *Partially Observable Markov Decision Process(POMDP)* problem. Recent research in the autonomous driving domain attempt to incorporate into the motion planning phase uncertainties induced by sensing noise [20], limited perception and occlusion [21], [22]. However, it is difficult to expand the POMDP framework to AV problems with continuous state and action space and solving the POMDP problem is still computationally expensive for real-world implementations.

*Sampling-based motion planning* has been recognized as effective tools for AV motion planning applications with complex dynamics and constraints. Extensions have been made to sampling-based frameworks (such as *Rapidly-exploring Random Trees*, RRT) to allow for the integration of uncertainties. For instance, the Chance-Constrained RRT (CC-RRT) [23] (and its extension CC-RRT* [24]) models the sensing uncertainties as Gaussian distributions and generate trees of risk-bounded trajectories. Similarly, the Rapidly-exploring Random Belief Tree (RRBT) in [25] incorporates propagation of motion and sensing uncertainties in the RRT* framework to ensure probabilistic feasibility.

Despite the appealing performance achieved, the aforementioned approaches are normally designed under the assumption of simplified, idealized model of sensing uncertainty and their effectiveness remains to be validated for systems with uncertainties of DNN-based components.

## III. PROBLEM FORMULATION

### A. Quantifying Uncertainties in DNN

Following the BDL paradigm, EU of DNNs can be estimated by placing prior distributions over the network's weights and estimate the variation of weights with respect to given training data. Denote $\mathbf{X} = \{\mathbf{x}_1...\mathbf{x}_n\}$ and $\mathbf{Y} = \{\mathbf{y}_1...\mathbf{y}_n\}$ as the training dataset and their corresponding ground truth, respectively. Denote a DNN as $f^{\mathbf{W}}(\cdot)$ with $L$ layers and weight parameters $\mathbf{W} = \{\mathbf{w}_i\}_{i=1}^{L}$ (where $\mathbf{w_i} = [w_{ij}]^T$ is the weight of the $j$th unit of the $i$th layer). In the BDL framework, $\mathbf{w}_i$ distributes according to $p(w)$. Given the training dataset $\{\mathbf{X}, \mathbf{Y}\}$, the posterior distribution of $\mathbf{W}$ is given by $p(\mathbf{W}|\mathbf{X}, \mathbf{Y})$. Denoting the input and expected prediction at testing time as $\mathbf{x}^*$ and $\mathbf{y}^*$, respectively, and the model likelihood as $p(\mathbf{y}^*|f^{\mathbf{W}}(\mathbf{x}^*))$. The predictive probabilistic output of the neural network at testing time is given by:

$$p(\mathbf{y}^*|\mathbf{x}^*, \mathbf{X}, \mathbf{Y}) = \int p(\mathbf{y}^*|f^{\mathbf{W}}(\mathbf{x}^*)p(\mathbf{W}|\mathbf{X}, \mathbf{Y})d\mathbf{w} \quad (1)$$

Since analytical inference of posterior distribution $p(\mathbf{W}|\mathbf{X}, \mathbf{Y})$ is intractable [10], it is feasible to use approximation inference technique to estimate a tractable posterior distribution $\hat{q}^{\theta}(\mathbf{W})$ (parameterized by $\theta$). It is proved in [10] that applying dropout before weighted layer of neural networks is mathematically equivalent to a Bayesian approximation of a BDL model. Therefore, uncertainties of DNN can be analytically evaluated via Monte Carlo (MC) dropout approximation, which operates by sampling from the approximate distribution ($\mathbf{W}_k \sim \hat{q}^{\theta}(\mathbf{W}), k \in \{1...N\}$) via multiple stochastic forward passes of same inputs. Taking advantage of MC dropout approximation, the predictive

variance of the DNN model for regression tasks can be estimated by:

$$\text{Var}_{q(\mathbf{y}^*|\mathbf{x}^*)}(\mathbf{y}^*) = \mathbb{E}_{q(\mathbf{y}^*|\mathbf{x}^*)}(\mathbf{y}^{*T}\mathbf{y}^*) \\ - \mathbb{E}_{q(\mathbf{y}^*|\mathbf{x}^*)}(\mathbf{y}^*)^T \mathbb{E}_{q(\mathbf{y}^*|\mathbf{x}^*)}(\mathbf{y}^*) \quad (2)$$

where:

$$\mathbb{E}_{q(\mathbf{y}^*|\mathbf{x}^*)}(\mathbf{y}^{*T}\mathbf{y}^*) \approx \sigma^2 + \frac{1}{T}\sum_{t=1}^{T} f^{\hat{\mathbf{W}}_t}(\mathbf{x}^*)^T f^{\hat{\mathbf{W}}_t}(\mathbf{x}^*) \quad (3)$$

$$\mathbb{E}_{q(\mathbf{y}^*|\mathbf{x}^*)}(\mathbf{y}^*) \approx \frac{1}{T}\sum_{t=1}^{T} f^{\hat{\mathbf{W}}_t}(\mathbf{x}^*) \quad (4)$$

and $T$ denotes the number of stochastic forward passes, the weights $\hat{\mathbf{W}}_t$ of each forward pass are sampled from the approximated distribution: $\hat{\mathbf{W}}_t \sim \hat{q}^\theta(\mathbf{W})$. $\sigma^2$ represents uncertainty caused by inherent noise in the observation data, which corresponds with AU.

The AU can be evaluated by estimating the aforementioned likelihood of observation $p(\mathbf{y}^*|f^{\mathbf{W}}(\mathbf{x}^*))$ via integration of loss function in the training of the DNN, which will be described in details in Section IV.

### B. Chance-Constrained Motion Planning

For AV applications involving uncertainties, outputs of perception module are the belief (i.e. estimated mean and variance) of surrounding traffic participants' states. Therefore, instead of using deterministic feasibility constraints, one of the major objectives of motion planning is to bound the probability of collision, and such chance-constrained problem can be formulated as:

**Problem 1. (chance-constrained motion planning)** Given an agent described by:

$$\mathbf{x}_{t+1} = f(\mathbf{x}_t, \mathbf{u}_t), \mathbf{x}_t \in \mathcal{X}, \mathbf{u}_t \in \mathcal{U} \quad (5)$$

(where $\mathbf{x}_t$ and $\mathbf{u}_t$ denote the state and control vectors at time $t$, respectively.) and given an initial state $\mathbf{x}_{init}$, the goal state $\mathbf{x}_{goal}$, the set of belief of detected obstacles in the environment $\mathcal{B}_t = \{B_i\}, (i = 1...N_{obs})$, generate a sequence of states and control inputs $\pi^* = \{\mathbf{x}_t, \mathbf{u}_t\}, (t \in [0, \Delta t])$ in $\Delta t$ that traverse from $\mathbf{x}_{init}$ to $\mathbf{x}_{goal}$, and minimize the cost of operation:

$$\pi^* = \arg\min_\pi \sum_{t}^{t+\Delta t} J(\pi_t, \mathcal{B}_t) \quad (6)$$

while satisfying the following chance-constraint:

$$P(\text{collision}|\pi^*, \mathcal{B}) = 1 - P\left(\bigwedge_t^{t+\Delta t} \neg\text{collision}|\pi^*, \mathcal{B}\right) \\ < \Delta \quad (7)$$

where $P(\text{collision})$ indicates the probability of collision with any of the obstacles in the environment. $\Delta$ is a predefined upper-bound on probability of collision(risk) during operation.

### C. Overall System Framework

In this work, we consider the typical configuration of an AV system, in which on-board sensing devices provide measurements of surrounding environments for the AV software stack, and its perception function uses such sensor readings to build an abstract model of the environment for the motion planning module to generate reference trajectories. The proposed framework mainly consists of an environment perception module and a motion planning module. The *environment perception module* takes as input the 3D point clouds of the surrounding environment captured by the vehicle-mounted LiDAR device, and it detects vehicles from point clouds and makes predictions of their 3D bounding boxes, along with the associated quantified uncertainties, using the proposed **Probabilistic PointRCNN** model. These classification and regression results as well as their uncertainties are then fed to the chance-constrained motion planning module based on the proposed **PU-RRT** algorithm. To explicitly incorporate the quantified perception uncertainties, a spatial representation of the detected object with multidimensional additive uncertainty is established to formulate the chance constraint. Based on the chance constraint, the PU-RRT checks probabilistic feasibility when expanding the tree of trajectories, and selects best and safe trajectories for execution at the same time. In this manner, the overall system generates trajectories that bound the risk under uncertainties. Details of the two primary components of the proposed framework are presented in the following section.

## IV. Proposed Methods

### A. Probabilistic PointRCNN

*1) Network Architecture:* The architecture of the proposed Probabilistic PointRCNN is depicted in Fig. 1. The two-stage PointRCNN is adopted as the backbone network, which consists of a 3D region proposal sub-network (RPN, stage 1) and a bounding box refinement sub-network (stage 2).Taking the 3D LiDAR point cloud as the input, the RPN generates region proposals of 3D bounding boxes of vehicles via point cloud segmentation. Point-wise feature vectors are also learned in this stage. In the stage 2 sub-network, the 3D points and their learned features from RPN are pooled and points within region proposals are selected, yielding the local point-wise features of each region proposal: $\mathbf{p} = [x_p, y_p, z_p, r_p, m_p, \mathbf{f}_p]$ ($x_p, y_p, z_p$: 3D coordinates, $r_p$: LiDAR intensity, $m_p$: segmentation mask, $\mathbf{f}_p$: feature vector from RPN). Such local features are concatenated with global semantic features $\mathbf{f}_b$ from RPN to generate rich features containing local and global information through another encoder. Outputs of the proposed network consist of classification of detected object $\{c_i\}$, along with their refined 3D bounding boxes: $\{\mathbf{b}_i\}(\mathbf{b}_i = (x_i, y_i, z_i, h_i, w_i, l_i, \theta_i)^T)$, as well as the quantified AU, EU and spatial representations of combined uncertainty.

Two major extensions are made to the original network: *intermediate layers* with MC dropout are introduced in stage 2 for the quantification pf EU, and *AU uncertainty layers* are added for modeling AU.

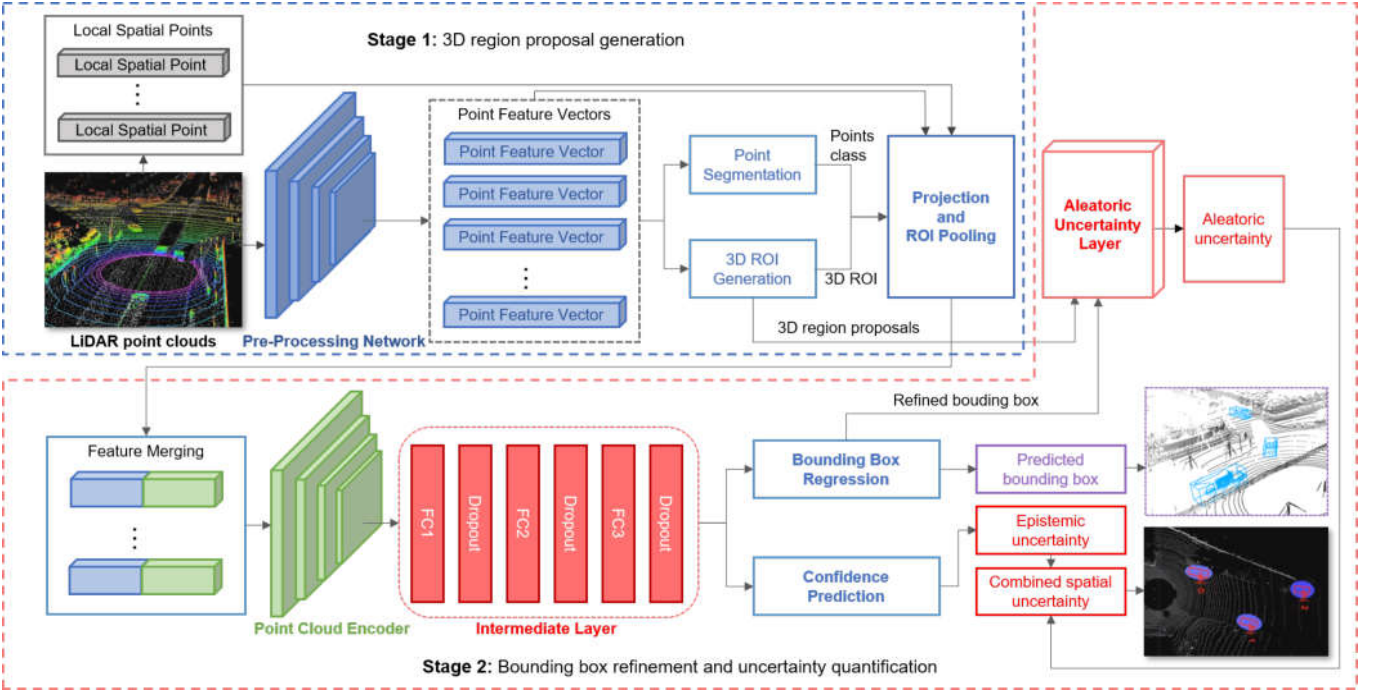

Fig. 1. The Bayesian PointRCNN architecture for 3D object detection and uncertainty quantification from LiDAR point clouds, developed based on the 2-stage PointRCNN backbone [26]. Uncertainty quantification-related extensions are marked in pink (ROI: region of interest; FC: fully-connected layer).

*2) Intermediate Dropout Layers:* Following the principle specified in Section III A, we utilize MC dropout technique as a Bayesian approximation for EU quantification. 3 fully-connected (FC) layers are introduced downstream the pooling and encoder of stage 2, where the first and second hidden layer contain 512 and 1024 neurons, respectively. The number of neurons of the last layer equals to the dimension of input of subsequent proposal refinement network. Each FC layer is followed by a dropout layer to perform the stochastic forward pass. For each input, the network perform $T$ times of forward passes, each with stochastic dropout. In this manner, the network draws samples from the weights' posterior distribution and the EU of regression task (i.e. 3D bounding boxes estimation) can be quantified following Eq. (3) and (4).

As for the classification EU, we use *Predictive Entropy* (PE) and *Mutual Information* (MI) [6] as metrics for quantification of the network's confidence in its classification results.

The Predictive Entropy is given by:

$$\mathbb{H}(\mathbf{y}^*|\mathbf{x}^*) = -\sum_c p(c|\mathbf{x}^*, \mathcal{D}) \log p(c|\mathbf{x}^*, \mathcal{D})$$
$$p(c|\mathbf{x}^*, \mathcal{D}) = \frac{1}{T} \sum_t^T p(c|\mathbf{x}^*, \hat{\mathbf{W}}_t) \quad (8)$$

where $\hat{\mathbf{W}}_t \sim q_{\boldsymbol{\omega}}(\boldsymbol{\theta})$ denotes the weight parameters of the $t$th forward pass, $p(c|\mathbf{x}^*, \hat{\boldsymbol{\omega}}_t)$ is the probability of an object being classified as $c$ at $t$, which is given by the softmax score in the proposed network (i.e. $p(c|\mathbf{x}^*, \hat{\boldsymbol{\omega}}_t) \approx \text{Softmax}(c, t)$). PE indicates the network's confidence in its predictions, and it reaches its maximum when the network is highly uncertain with its predictions (i.e. $p(c)$ is close to 0.5). The Mutual Information is given by:

$$\mathbb{I}(\mathbf{y}^*, \boldsymbol{\omega}|\mathbf{x}^*, \mathcal{D}) = \mathbb{H}(\mathbf{y}^*|\mathbf{x}^*) - \mathbb{E}_{p(\boldsymbol{\omega}|\mathcal{D})}(\mathbb{H}(\mathbf{y}^*|\mathbf{x}^*))$$
$$\mathbb{E}_{p(\boldsymbol{\omega}|\mathcal{D})}(\mathbb{H}(\mathbf{y}^*|\mathbf{x}^*)) = \\ -\frac{1}{T}\sum_t \sum_c p(c|\mathbf{x}^*, \hat{\mathbf{W}}_t) \log p(c|\mathbf{x}^*, \hat{\mathbf{W}}_t) \quad (9)$$

MI measures the network's confidence in its predictions in repetitive trials, and it reaches its peak when the predictions from multiple forward passes deviate from each other. Therefore, PE and MI can be used to filter out mis-detections in the decision-making and planning process.

*3) Aleatoric Uncertainty Layers:* The aleatoric uncertainty corresponds to the inherent noise in the observations. To capture the AU in the regression task, the network needs to estimate the observation likelihood $p\left(\mathbf{y}^*|f^\mathbf{W}(\mathbf{x}^*)\right)$ as in Eq.(1). As the elements of the 3D bounding box vector are mutually independent, one can model the likelihood as multi-variate Gaussian distribution with diagonal covariance matrix [14]:

$$p(\mathbf{y}^*|\mathbf{x}^*, \boldsymbol{\omega}) \sim \mathcal{N}(f^{\hat{\mathbf{W}}_t}, \boldsymbol{\Sigma}(\mathbf{x}^*))$$
$$\mathbf{Var}^{\text{alea}}(\mathbf{x}^*) = diag(\boldsymbol{\sigma}_{\mathbf{x}^*}^2) \quad (10)$$

In the regression task, the variance parameters are encoded as $\hat{\boldsymbol{\sigma}}_{\mathbf{x}^*}^2 = [\sigma_x^2, \sigma_y^2, \sigma_z^2, \sigma_h^2, \sigma_w^2, \sigma_l^2, \sigma_\theta^2]^\mathrm{T}$, of which each element corresponds to the uncertainty of an element in the predicted bounding box vector.

To estimate the AU, we introduce AU quantification layers to the network (Fig. 2). As the refined bounding boxes in stage 2 are generated from the candidate RPs from the stage 1 sub-network, the uncertainties come from the two stages and we use both the feature vectors of bounding boxes and corresponding RPs as the input of FC AU layers to obtain $\lambda_{roi}$ and $\lambda_{pred}$. These processed two sources of features are then merged and fed to final fully-connected layers to obtain the estimated observation variance $\boldsymbol{\lambda_{x^*}}$. During the training phase, we use the following loss function for learning the AU of regression [14]:

$$\mathcal{L}(\mathbf{W}) = \frac{1}{D}\sum_d^D \frac{1}{2}\exp(-\lambda_d)\Delta^2 + \frac{1}{2}\lambda_d \quad (11)$$

$\lambda_d$ represents individual element in the $\boldsymbol{\lambda}$ of each sample in the training dataset. $\Delta$ represents the residual loss of each element in the predicted vector. For numerical stability, we use logarithmic term ($\boldsymbol{\lambda_{x^*}} := \log \boldsymbol{\sigma^2_{x^*}}$) to avoid possible division by zero. In this manner the observation noise-related uncertainty can be learned from the input data.

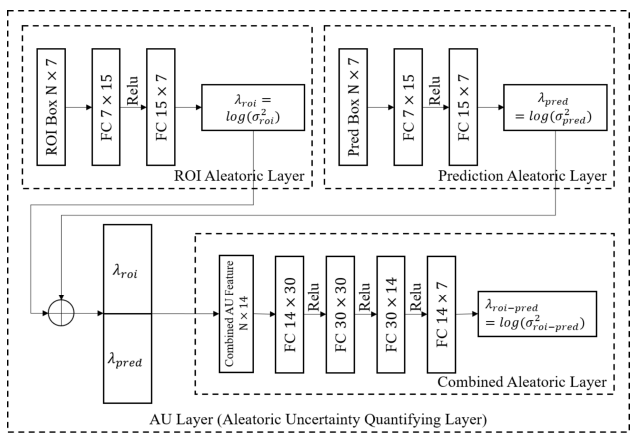

Fig. 2. Aleatoric uncertainty quantification layers.

### B. Spatial formulation of Combined Perception Uncertainties

In order to propagate and explicitly incorporate DNN perception uncertainties in the subsequent motion planning phase, we combine the AU and EU and derive a spatial formulation of the uncertainty in the planning space. Following the BDL principle as in Eq. (2), the combined uncertainty for each predicted bounding box can be approximated by:

$$\mathbf{Var}(\mathbf{y}_i^*) = \frac{1}{T}\sum_t^T \hat{\mathbf{y}}_i\hat{\mathbf{y}}_i^\mathrm{T} + \left(\frac{1}{T}\sum_t^T \hat{\mathbf{y}}_i\right)\left(\frac{1}{T}\sum_t^T \hat{\mathbf{y}}_i\right)^\mathrm{T} \\ + \frac{1}{T}\sum_t^T \hat{\boldsymbol{\sigma}}_t^2 \quad (12)$$

where the first two terms relate to the EU and the last term is the AU. To incorporate the modeled uncertainty into the motion planning space, we extracted the elements related to position, scale and orientation $[\sigma^2_{x,i}, \sigma^2_{y,i}, \sigma^2_{w,i}, \sigma^2_{l,i}, \sigma^2_{\theta,i}]^\mathrm{T}$ from the predicted variance vector (Eq. (12)). As elements in the predicted bounding box vector are mutually-independent variables with Gaussian distribution, the variances of elements related to a certain dimension (e.g. lateral and longitudinal) are additive. Therefore, we formulate the quantified uncertainties in the lateral and longitudinal dimensions (i.e. $X$ and $Y$ of the vehicle body-fixed frame) of the estimated bounding box. For the fist step, the variance of estimation width and length ($\sigma_w^2, \sigma_l^2$) are added to those of x-position and y-position ($\sigma_x^2, \sigma_y^2$) to derive the total uncertainty in the lateral and longitudinal dimension, respectively:

$$\begin{aligned}\sigma_{lat} &= \sqrt{\sigma_x^2 + \sigma_w^2} \\ \sigma_{lon} &= \sqrt{\sigma_y^2 + \sigma_l^2}\end{aligned} \quad (13)$$

For simplicity, we further incorporate the uncertainty in the estimated orientation ($\sigma_\theta$) by projecting the estimated width and length based on $\sigma_\theta$ and deriving orientation-related additive variances ($\Delta_a, \Delta_b$):

$$\begin{aligned}\Delta_a &= \frac{l}{2}\left(1 - w\sqrt{\frac{(1 + \tan^2(\sigma_\theta))}{(w^2 + l^2\tan^2(\sigma_\theta))}}\right) \\ \Delta_b &= \frac{w}{l}\Delta_a\end{aligned} \quad (14)$$

Therefore, by incorporating the lateral and longitudinal uncertainties into the predicted bounding boxes, we derive a ellipse-like formulation of the detected object, with the ultimate scale of the two dimensions given by:

$$\begin{aligned}L_a &= l/2 + \sigma_{lon} + \Delta_a \\ L_b &= w/2 + \sigma_{lat} + \Delta_b\end{aligned} \quad (15)$$

Examples of detected objects with additive perceptions uncertainties are depicted in Fig. 3.

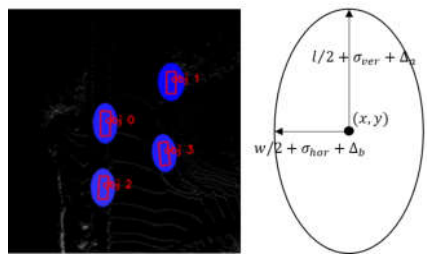

Fig. 3. Examples of representation of bounding boxes of detected vehicles with additive quantified perception uncertainties.

### C. Uncertainty-aware Motion Planning

To bound the operational risk by considering perception uncertainties of DNN, we propose an extension of the real time Closed-loop RRT (CL-RRT) [27], namely PU-RRT, which explicitly incorporates perception and motion uncertainties by checking the probability of collision based on chance constraints. We firstly formulate the chance constraint based on the quantified uncertainty and adopt a computational-efficient risk evaluation approach [23].

*1) Chance-Constraint formulation and risk evaluation:*
The vehicle system can be modeled as a linear time-invariant system with additive Gaussian process noise [23]:

$$\mathbf{x}_{t+1} = \mathbf{A}\mathbf{x}_t + \mathbf{B}\mathbf{u}_t + \boldsymbol{\gamma}_t$$
$$\mathbf{x}_0 \sim \mathcal{N}(\hat{\mathbf{x}}_0, \boldsymbol{\Sigma}_0) \quad (16)$$
$$\boldsymbol{\gamma}_t \sim \mathcal{N}(0, \boldsymbol{\Sigma}_\gamma)$$

where $\mathbf{x}_t$ is the vehicle's state at time $t$, $\mathbf{u}_t$ is the control input vector and $\boldsymbol{\gamma}_t$ is the noise vector which models the external disturbance and inherent system noise. $\mathcal{N}(0, \boldsymbol{\Sigma}_\gamma)$ denotes Gaussian distribution with zero-mean and variance $\boldsymbol{\Sigma}_\gamma$. $\mathcal{N}(\hat{\mathbf{x}}_0, \boldsymbol{\Sigma}_0)$ denotes the belief of the initial state of vehicle.

To formulate the system uncertainty as chance constraint formulation for probabilistic feasibility evaluation, we propagate the system uncertainty over time based on the system model (Eq. (16)). Given the control sequence of time step $t \in \{0, 1, ..., t_f\}$, the distribution of the state $\mathbf{x}_t$, the mean $\hat{\mathbf{x}}_t$ and covariance $\boldsymbol{\Sigma}_t$ can be propagated based on the estimation of the previous time step:

$$\hat{\mathbf{x}}_{t+1} = \mathbf{A}\hat{\mathbf{x}}_t + \mathbf{B}\mathbf{u}_t$$
$$\boldsymbol{\Sigma}_{t+1} = \mathbf{A}\boldsymbol{\Sigma}_t \mathbf{A}^T + \boldsymbol{\Sigma}_\gamma \quad (17)$$
$$\forall t \in \{0, 1, ..., t_f - 1\}$$

As stated in Eq. (7), the chance constraint requires that the probability of collision with any of surrounding obstacles is bounded within safe threshold ($\Delta = 1 - p_{safe}$). Denote the number of obstacles detected by the perception module as $K$, and $z_j$ represents the coordinates of the geometric center of the $j$th obstacle's bounding box. We model the ego vehicle as a convex polygon defined by conjunction four linear constraints ($\bigwedge_{i=1,2,3,4} a_i^T x_t < b_i$, Fig. 4). the probability of the ego vehicle colliding with $j$th obstacle is given by:

$$P_{\text{collision},j} = P\left(\bigwedge_{i=1,2,3,4} \mathbf{a}_i^T \mathbf{z}_j < \mathbf{b}_i\right) \quad (18)$$

Therefore, the probability that any of the linear constraints in eq. (18) being satisfied is an upper bound on the probability of collision:

$$P_{\text{collision},j} = P\left(\bigcap_{i=1,2,3,4} \mathbf{a}_i^T \mathbf{z}_j < \mathbf{b}_i\right)$$
$$\leq P\left(\mathbf{a}_i^T \mathbf{z}_j < \mathbf{b}_i\right) \quad (19)$$
$$\forall j \in \{1, ..., K\}$$

Recall that the covariance of the location of the obstacle's geometric center is given by $\boldsymbol{\Sigma}_j$ ($\boldsymbol{\Sigma}_j = [\sigma_{lat} + \Delta_b, \sigma_{lon} + \Delta_a]T$, as in Eq. (13) and (14)) and the covariance of the ego vehicle's location is $\boldsymbol{\Sigma}_{x_t}$ (propagated as in Eq. (17)). By integrating these uncertainties, the probability that the $i$th linear constraint is satisfied by the $j$th obstacle can be calculated as:

$$\Delta_{ijt} = P\left(\mathbf{a}_i^T \mathbf{z}_j - \mathbf{b}_i < 0\right)$$
$$= \frac{1}{2}\left(1 - erf\left(\frac{\mathbf{a}_i^T \mathbf{z}_j - \mathbf{b}_i}{\sqrt{2\mathbf{a}_i^T (\boldsymbol{\Sigma}_j + \boldsymbol{\Sigma}_{x_t}) \mathbf{a}_i}}\right)\right) \quad (20)$$

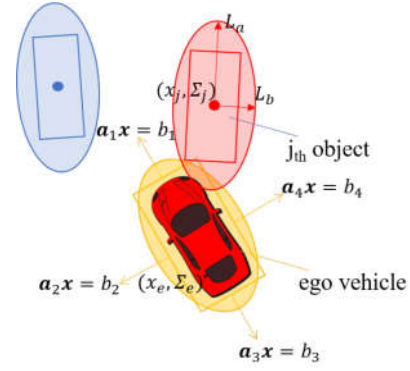

Fig. 4. Spatial representation of DNN perception uncertainties and evaluation of risk of collision.

where $\Delta_{ijt}$ denote the probability that the $j$th obstacle satisfies $i$th linear constraint of the ego-vehicle at time step $t$, and $erf(\cdot)$ is the Gauss error function which can be solved quickly using pre-calculated look-up tables.

To simplify the calculation, as Eq. (19) specifies the upper bound of the probability of collision, the actual probability of vehicle colliding with $j$th obstacle does not exceed the minimum probability of satisfying any of the four linear constraints:

$$P_{\text{collision},j} \leq \min_{i=1,2,3,4} P\left(\mathbf{a}_i^T \mathbf{z}_j < \mathbf{b}_i\right) \quad (21)$$

Then for all obstacles, the overall probability of collision at time $t$ can be approximated as:

$$P_{\text{collision}}(t) \leq \sum_j^K \min_{i=1,2,3,4} P\left(\mathbf{a}_i^T \mathbf{z}_j < \mathbf{b}_i\right)$$
$$\approx \sum_j^K \min_{i=1,2,3,4} \Delta_{ijt} \quad (22)$$

Denote $p_{safe}$ as the pre-defined safety threshold during operation, the overall probability of collision (Eq. (22)) should satisfy the following constraint:

$$P_{collision}(t) = \Delta_t < \Delta = 1 - p_{safe} \quad (23)$$

*2) PU-RRT:* Similar to the real-time CL-RRT [27], the PU-RRT algorithm mainly consists of two operations: the Tree Expansion (**Algorithm 1**) and the Execution (**Algorithm 2**). The former operates by growing a tree of dynamically and probabilistic feasible trajectories via random sampling, while the latter simultaneously selects the current best trajectory for execution by the controller. Given the quantified perception uncertainty and chance constraint formulated by Eq. (20 -23), we make extensions to both operations to enable the algorithm to explicitly check if the collision of probability of nodes and trajectories satisfies chance constraints.

Details of the Tree Expansion phase is presented in Algorithm 1. The algorithm takes the initial and goal state, as well as the current tree as the initial input. At the beginning of each

**Algorithm 1: PU-RRT, Tree Expansion**

**Input:** vehicle state $(\mathbf{x}_0, \mathbf{\Sigma}_0)$, goal $\mathbf{x}_{goal}$, detected obstacles with uncertainty $b_i, Q_i$, configuration space $C$, safety threshold $p_{safe}$, search tree $T$
**Output:** Tree $T$

1. Update the environment model using $b_i, Q_i$;
2. Sampling $\mathbf{x}_{sample}$ from the configuration space $C$, filter out mis-detections based on MI and PE;
3. Sort the nodes in the tree using heuristics;
4. Choose $M$ nodes according to *step*;
5. **for** *each node $n_i, i \in \{1, ..., M\}$* **do**
6.     Get the position $\mathbf{x}_t$ and uncertainty matrix $\mathbf{\Sigma}_t$;
7.     Calculate chance-constrained risk
8.     $\Delta_t \leftarrow \text{getCCRisk}(\mathbf{x}_t, \mathbf{\Sigma}_t, Q)$;
9.     $k = 0$;
10.     **while** $\mathbf{x}_{t+k|t}$ *has not reached* $\mathbf{x}_{sample}$ **do**
11.         Steering $\mathbf{x}_{t+k+1|t} \leftarrow \text{steer}(\mathbf{x}_{t+k|t}, \mathbf{x}_{sample})$;
12.         Uncertainty propagation
13.         $\mathbf{\Sigma}_{t+k+1|t} \leftarrow \text{probPropagation}(\mathbf{x}_{t+k|t}, \mathbf{\Sigma}_{t+k|t})$;
14.         $\Delta_{t+k+1|t} \leftarrow \text{getCCRisk}(\mathbf{x}_{t+k+1|t}, \mathbf{\Sigma}_{t+k+1|t}, Q)$;
15.         **if** *isFeasible*$(\mathbf{x}_{t+k+1|t})$ *and* $\Delta_{t+k+1|t} < 1 - p_{safe}$ **then**
16.             Generate new tree node $n_{t+k+1|t} \leftarrow \text{generateNode}(\mathbf{x}_{t+k+1|t}, \mathbf{\Sigma}_{t+k+1|t})$;
17.             Add node $n_{t+k+1|t}$ to tree $T$;
18.             $k \leftarrow k + 1$;
19.         **else**
20.             Exit while;
21.         **end**
22.     **end**
23.     **for** *generated node $n$ by step* **do**
24.         Get the position of node $n$;
25.         Try connecting $n$ to $\mathbf{x}_{goal}$ (lines 10-22);
26.         **if** *connection to $\mathbf{x}_{goal}$* **then**
27.             Store the path;
28.             Add generated nodes to Tree $T$;
29.             Update $C_{UB}$ value of all nodes in path;
30.         **end**
31.     **end**
32. **end**

**Algorithm 2: PU-RRT, Trajectory Execution**

**Input:** vehicle $\mathbf{x}_0$, goal $\mathbf{x}_{goal}$, safety threshold $p_{safe}$
**Output:** Reference path to controller

1. **while** *vehicle not reaches goal* **do**
2.     Initialize the uncertainty matrix $\mathbf{\Sigma}_0$ of ego-vehicle;
3.     Initialize tree $T \leftarrow \text{treeInitialize}(\mathbf{x}_0, \mathbf{\Sigma}_0)$;
4.     Update configuration space $C$;
5.     Update detection results $Q$;
6.     **while** *time limit $\Delta_t$ is not reached* **do**
7.         Expand the tree (Algorithm 1);
8.     **end**
9.     Select best path (Algorithm 3);
10.     Apply the best path;
11. **end**

cycle, the algorithm updates environment model based on the detected bounding boxes with quantified uncertainty from the perception module (line 1), and mis-detections are filtered out based on the classification uncertainty (MI and PE). Algorithm 1 expands the search tree continuously via randomly sampling nodes from the configuration space until the time for expansion reaches maximum. We incorporate the quantified uncertainties into the nearest node selection step (line 4) and the feasibility checking step (line 15).

To find the nearest nodes to be connected to the newly-sampled node $\mathbf{x}_{sample}$, the cost for connection is evaluated based on two metrics: the distance for steering to $\mathbf{x}_{sample}$ (*dist*), and the probability of collision (i.e. the risk, $\Delta_{it}$). Therefore, the overall cost of node is given by:

$$C(n_i) = k_{cc}\Delta_{it} + k_{dist}dist(n_i, n_{start}) \quad (24)$$

where $k_{cc}$ and $k_{dis}$ are the weights for adjusting the importance of two metrics. In addition, before connecting new nodes to the existing tree, the algorithm generates closed-loop trajectory that steers the vehicle to $\mathbf{x}_{sample}$, and checks the risk of the simulated trajectory (line 15). Only the probabilistic feasible portion of the trajectory is kept and new nodes are generated and connected to the existing tree (line 16 and 17). In this manner, the algorithm attempts to connect newly-sampled nodes to the tree nodes with low risk and operation cost.

The trajectory execution phase is described in **Algorithm 2**. To select the current optimal path, nodes are evaluated based on their costs. Each generated node keeps a lower bound $C_{LB}$ and an upper bound cost-to-go value $C_{UB}$, which serve as metrics for the evaluation the cost from the specified node to the goal. $C_{LB}$ is determined by the direct Euclidean distance between the node and the goal. When a feasible trajectory from the node to the goal is available, $C_{UB}$ is updated by propagating the cost metrics from goal backwards to the specified node [27] to check if there exists a path with lower upper bound cost-to-go. Therefore, $C_{UB}$ is given by:

$$C_{UB} = \begin{cases} +\infty & \text{no trajectory to the goal} \\ \min_c(e_c + C_{UB_c}) & \text{trajectory to the goal exists} \\ C_{LB} & \text{node inside goal region} \end{cases} \quad (25)$$

where $c$ represents the index of child nodes, $e_c$ is the cost from the node to the child node $c$, and $C_{UB_c}$ represents the upper bound cost at node $c$. $e_c$ is calculated in a similar manner as in (24) to incorporate the risk of trajectories in the path selection:

$$e_c(n_i) = k_{cc}\Delta_{it} + k_{dis}dis(n_i, n_c) \quad (26)$$

These cost-to-go values are used in trajectory selection in the **Algorithm 3**. In this manner, the probabilistic feasibility-related cost metric allows for the selection of optimal path that bounds the risk while lowering the operation cost.

```
Algorithm 3: PU-RRT, Path Selection
  Input: Feasible paths, Searching Tree T
  Output: Reference path
1 if the number of feasible path ≠ 0 then
2     for each feasible path p_i do
3         Get C_UB of each node;
4         Record max C_UBi of the node in path p_i;
5     end
6     Choose the path with lowest C_UBi;
7 else
8     No feasible path;
9     path← findCloseToGoal(T, x_goal);
10 end
```

## V. EXPERIMENTAL EVALUATION

### A. Experiment Setup

*1) Network Training and Validation:* Our proposed Probabilistic PointRCNN is trained using a platform with 2 TESLA V100 GPUs. The KITTI dataset [28] combined with simulated LiDAR data collected from CARLA simulator are used as the training and validation dataset. During training phase, We fix the dropout rate of intermediate layer to 0.5 and the times of MC Dropout sampling is set to 25. In the training phase. The RPN is trained with 200 epochs, while the RCNN is trained firstly with regular loss function by 40 epochs and then trained with AU-related loss function (11) by 120 epochs. The adopted optimization method is adam-onecycle with weight decay 0.001 and momentum value 0.9. The initial learning rate is 0.002 and decays by 0.5 every 40 epochs. The times of MC Dropout sampling during inference is set to 6 in simulation. The simulation test is performed using one GTX 1070 GPU.

*2) Parameters of PU-RRT:* For PU-RRT, we set max iteration time to 200 for tree expansion. For nearest node selection, the number of candidate nodes $M$ is set to 15, and the time interval for tree expansion is 3s.

*3) Scenario Setup:* We implement the proposed Probabilistic PointRCNN and PU-RRT as ROS (Robotics Operation System) nodes interacting with other function modules. The proposed approaches are validated in simulated scenarios built using CARLA simulator (Fig 5). The designed scenario is a multi-lane road with dense traffic where the ego vehicle must perceive the surrounding obstacles and plan safe trajectories to perform lane-changing and reach the goal position. The scenario involves factors that may induce uncertainty in DNN: variation of observation distance and angle, as well as occlusion. The simulated LiDAR measurements from CARLA are fed to the Probabilistic PointRCNN node for detection, and the PU-RRT node takes as input the detected bounding boxes and quantified uncertainty to generate trajectories.

### B. Experiment Results

*1) Object Detection and Uncertainty Quantification Results:* To analyze how different factors affect the uncertainty of DNN, we evaluate the proposed Probabilistic PointRCNN by detecting objects at varying distances and azimuth of observation. Object detection results are depicted in Fig. 6 along with plotted quantified uncertainty. Uncertainties in the estimated location, lateral and longitudinal scale increase as the detected vehicle moves away from the ego vehicle (Fig. 6(a)). In addition, undesirable azimuth of observation also result in increased uncertainties. These results indicate that the proposed approach can effectively quantify uncertainty of DNN and capture the effects of various factors.

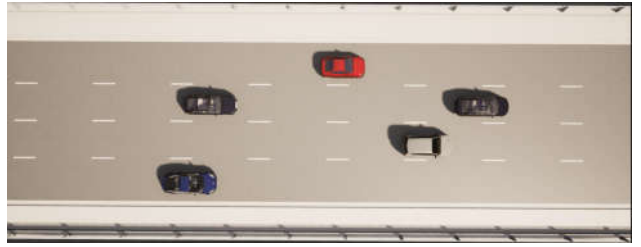

Fig. 5. Simulated multi-lane changing scenario

*2) Motion Planning Results:* To compare the performance of the proposed motion planner with that of existing approaches. We test the performance of the following three motion planners:

**PU-RRT**: the proposed motion planning algorithm;

**CC-RRT**: conventional chance-constrained RRT planner which assumes a fixed scale of uncertainty in the detected obstacles instead of explicitly estimating the uncertainties of DNN.

**CL-RRT**: closed-loop RRT that does not account for any uncertainty.

Examples of planned trajectories generated the three planners in the simulated scenario are depicted in Fig 7. Comparisons of performance of three planners are shown in Table I. As can be seen from the results, the proposed PU-RRT outperforms CL-RRT and in terms of trip success rate (ratio of trips that do not violate chance constraint) and risk (Risk-max, Risk-avg) due to the integration of perception uncertainties. In addition, compared with conservative CC-RRT, the PU-RRT can generate shorter trajectories (length of traj).

## VI. CONCLUSION

This paper presents a safe motion planning framework that allow for the integration of perception and motion uncertainties. A Bayesian DNN-based object detector is proposed to quantify the uncertainty of perception. By explicitly incorporating the quantified perception uncertainty and motion uncertainty, the prpoposed PU-RRT can bound the operation risk while achieving desirable efficiency.

Our future work will focus on more comprehensive validation of the proposed framework in more scenarios. In addition, the real-time performance of the proposed approaches will also be evaluated.

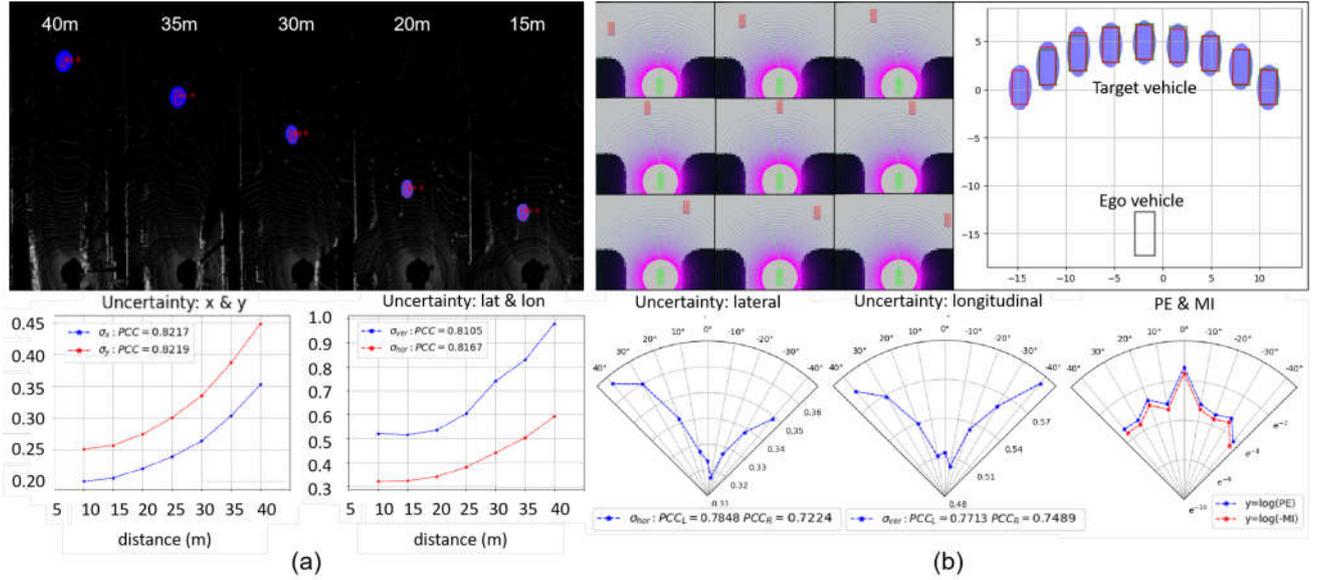

Fig. 6. Object detection and uncertainty quantification: (a) object detection at varying distances; (b) object detection at varying azimuth of observation

TABLE I
PERFORMANCE COMPARISON OF MOTION PLANNING ALGORITHMS

| $t_1$ | $Rate_{succ,1}$ | $Rate_{succ,2}$ | $Rate_{succ,3}$ | Risk-max | Risk-avg | $n_{waypoints}$ | Length of traj(m) |
|---|---|---|---|---|---|---|---|
| PU-RRT | 100% | 82% | 82% | 0.0061 | 0.0016 | 35.73 | 45.36 |
| CC-RRT | 100% | 67% | 67% | 0.0014 | 0.00022 | 40.14 | 49.75 |
| CL-RRT | 72% | 72% | 12% | 0.0390 | 0.014 | 28.94 | 39.70 |
| $t_2$ | $Rate_{succ,1}$ | $Rate_{succ,2}$ | $Rate_{succ,3}$ | Risk-max | Risk-avg | $n_{waypoints}$ | Length of traj(m) |
| PU-RRT | 100% | 100% | 100% | 0.0080 | 0.0027 | 25.57 | 35.56 |
| CC-RRT | 100% | 56% | 56% | 0.0055 | 0.00069 | 30.33 | 38.00 |
| CL-RRT | 81% | 81% | 33% | 0.016 | 0.0046 | 24.70 | 35.34 |
| $t_3$ | $Rate_{succ,1}$ | $Rate_{succ,2}$ | $Rate_{succ,3}$ | Risk-max | Risk-avg | $n_{waypoints}$ | Length of traj(m) |
| PU-RRT | 100% | 100% | 100% | 0.0053 | 0.0012 | 26.49 | 36.88 |
| CC-RRT | 100% | 91% | 91% | 0.0034 | 0.00052 | 28.24 | 37.67 |
| CL-RRT | 90% | 90% | 38% | 0.010 | 0.0024 | 26.25 | 36.31 |
| $t_4$ | $Rate_{succ,1}$ | $Rate_{succ,2}$ | $Rate_{succ,3}$ | Risk-max | Risk-avg | $n_{waypoints}$ | Length of traj(m) |
| PU-RRT | 100% | 99% | 99% | 0.0052 | 0.0011 | 24.85 | 33.74 |
| CC-RRT | 100% | 76% | 76% | 0.0010 | 0.00018 | 25.42 | 34.04 |
| CL-RRT | 93% | 93% | 55% | 0.016 | 0.0037 | 24.35 | 33.71 |

REFERENCES

[1] B. Paden, M. Čáp, S. Z. Yong, D. Yershov, and E. Frazzoli, "A survey of motion planning and control techniques for self-driving urban vehicles," *IEEE Transactions on intelligent vehicles*, vol. 1, no. 1, pp. 33–55, 2016.

[2] S. Grigorescu, B. Trasnea, T. Cocias, and G. Macesanu, "A survey of deep learning techniques for autonomous driving," *Journal of Field Robotics*, vol. 37, no. 3, pp. 362–386, 2020.

[3] M. Treml, J. Arjona-Medina, T. Unterthiner, R. Durgesh, F. Friedmann, P. Schuberth, A. Mayr, M. Heusel, M. Hofmarcher, M. Widrich *et al.*, "Speeding up semantic segmentation for autonomous driving," in *ML-ITS, NIPS Workshop*, vol. 2, no. 7, 2016.

[4] S. Zhang, R. Benenson, M. Omran, J. Hosang, and B. Schiele, "Towards reaching human performance in pedestrian detection," *IEEE transactions on pattern analysis and machine intelligence*, vol. 40, no. 4, pp. 973–986, 2017.

[5] E. Hüllermeier and W. Waegeman, "Aleatoric and epistemic uncertainty in machine learning: A tutorial introduction," *arXiv preprint arXiv:1910.09457*, vol. 5, 2019.

[6] Y. Gal, "Uncertainty in deep learning," *University of Cambridge*, vol. 1, no. 3, p. 4, 2016.

[7] A. Graves, "Practical variational inference for neural networks," in *Advances in Neural Information Processing Systems*, J. Shawe-Taylor, R. Zemel, P. Bartlett, F. Pereira, and K. Q. Weinberger, Eds., vol. 24. Curran Associates, Inc., 2011, pp. 2348–2356.

[8] C. Blundell, J. Cornebise, K. Kavukcuoglu, and D. Wierstra, "Weight uncertainty in neural network," in *International Conference on Machine Learning*. PMLR, 2015, pp. 1613–1622.

[9] A. Kucukelbir, D. Tran, R. Ranganath, A. Gelman, and D. M. Blei, "Automatic differentiation variational inference," *The Journal of Machine Learning Research*, vol. 18, no. 1, pp. 430–474, 2017.

[10] Y. Gal and Z. Ghahramani, "Dropout as a bayesian approximation: Representing model uncertainty in deep learning," in *international conference on machine learning*, 2016, pp. 1050–1059.

[11] ——, "Bayesian convolutional neural networks with Bernoulli approximate variational inference," in *4th International Conference on Learning Representations (ICLR) workshop track*, 2016.

[12] B. Lakshminarayanan, A. Pritzel, and C. Blundell, "Simple and scalable predictive uncertainty estimation using deep ensembles," in *Proceedings of the 31st International Conference on Neural Information Processing Systems*, ser. NIPS'17. Red Hook, NY, USA: Curran Associates Inc., 2017, p. 6405–6416.

[13] M. T. Le, F. Diehl, T. Brunner, and A. Knol, "Uncertainty estimation for deep neural object detectors in safety-critical applications," in *2018 21st International Conference on Intelligent Transportation Systems (ITSC)*, 2018, pp. 3873–3878.

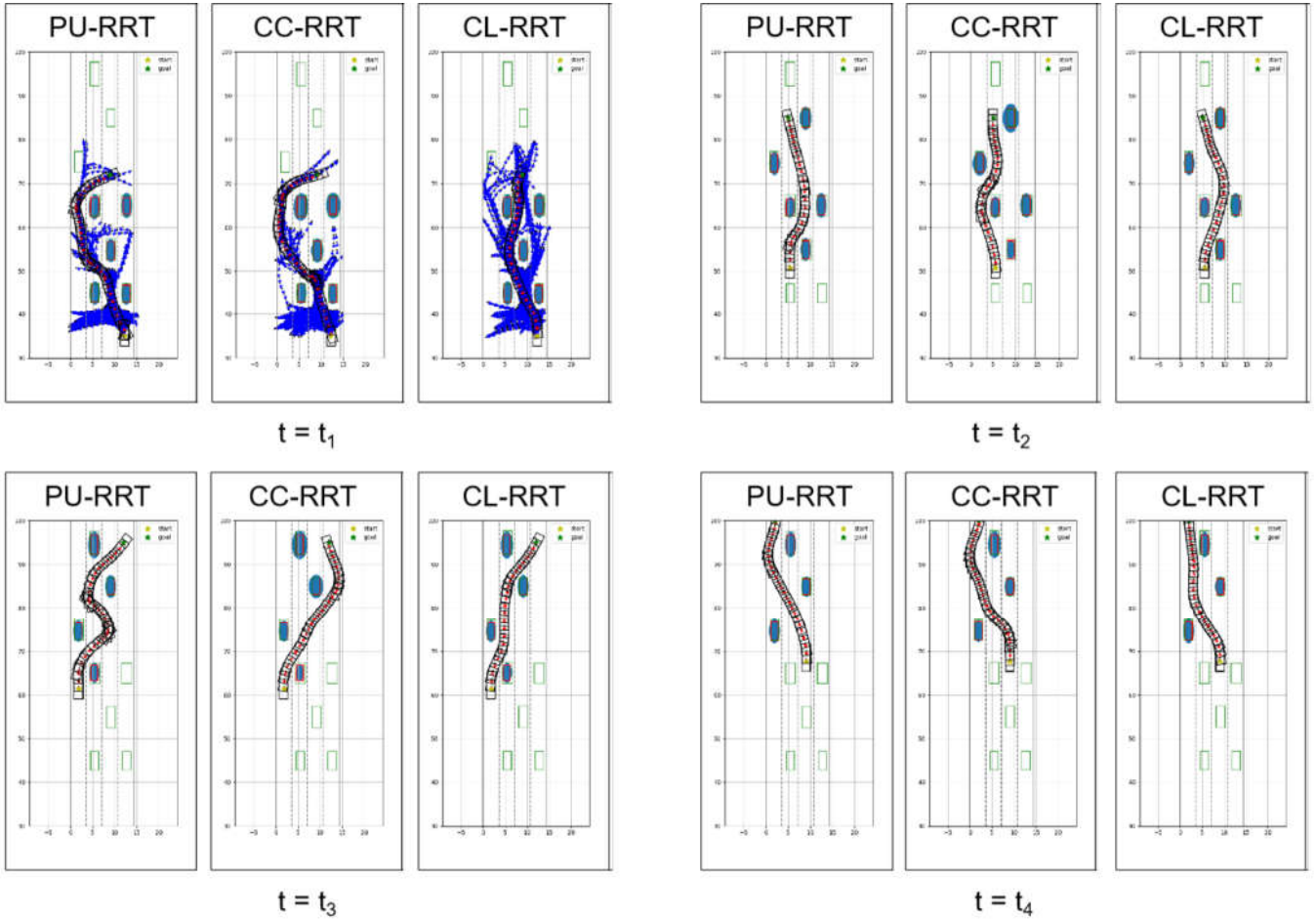

Fig. 7. Examples of trajectories generated by PU-RRT, CC-RRT and CL-RRT. Green/red rectangles: ground truth/estimated bounding boxes of surrounding vehicles, blue ellipses: spatial representation of uncertainties of estimated bounding boxes, black rectangles: planned trajectories of the ego vehicle

[14] A. Kendall and Y. Gal, "What uncertainties do we need in bayesian deep learning for computer vision?" in *Advances in neural information processing systems*, 2017, pp. 5574–5584.

[15] S. Choi, K. Lee, S. Lim, and S. Oh, "Uncertainty-aware learning from demonstration using mixture density networks with sampling-free variance modeling," in *2018 IEEE International Conference on Robotics and Automation (ICRA)*, 2018, pp. 6915–6922.

[16] M. Abdar, F. Pourpanah, S. Hussain, D. Rezazadegan, L. Liu, M. Ghavamzadeh, P. Fieguth, A. Khosravi, U. R. Acharya, V. Makarenkov *et al.*, "A review of uncertainty quantification in deep learning: Techniques, applications and challenges," *arXiv preprint arXiv:2011.06225*, 2020.

[17] D. Feng, L. Rosenbaum, and K. Dietmayer, "Towards safe autonomous driving: Capture uncertainty in the deep neural network for lidar 3d vehicle detection," in *2018 21st International Conference on Intelligent Transportation Systems (ITSC)*. IEEE, 2018, pp. 3266–3273.

[18] D. Feng, L. Rosenbaum, F. Timm, and K. Dietmayer, "Leveraging heteroscedastic aleatoric uncertainties for robust real-time lidar 3d object detection," in *2019 IEEE Intelligent Vehicles Symposium (IV)*. IEEE, 2019, pp. 1280–1287.

[19] G. P. Meyer and N. Thakurdesai, "Learning an uncertainty-aware object detector for autonomous driving," *CoRR*, vol. abs/1910.11375, 2019. [Online]. Available: http://arxiv.org/abs/1910.11375

[20] C. Hubmann, J. Schulz, M. Becker, D. Althoff, and C. Stiller, "Automated driving in uncertain environments: Planning with interaction and uncertain maneuver prediction," *IEEE Transactions on Intelligent Vehicles*, vol. 3, no. 1, pp. 5–17, 2018.

[21] C. Hubmann, N. Quetschlich, J. Schulz, J. Bernhard, D. Althoff, and C. Stiller, "A pomdp maneuver planner for occlusions in urban scenarios," in *2019 IEEE Intelligent Vehicles Symposium (IV)*, 2019, pp. 2172–2179.

[22] S. Brechtel, T. Gindele, and R. Dillmann, "Probabilistic decision-making under uncertainty for autonomous driving using continuous pomdps," in *17th International IEEE Conference on Intelligent Transportation Systems (ITSC)*, 2014, pp. 392–399.

[23] B. Luders, M. Kothari, and J. How, "Chance constrained rrt for probabilistic robustness to environmental uncertainty," in *AIAA guidance, navigation, and control conference*, 2010, p. 8160.

[24] W. Liu and M. H. Ang, "Incremental sampling-based algorithm for risk-aware planning under motion uncertainty," in *2014 IEEE International Conference on Robotics and Automation (ICRA)*, 2014, pp. 2051–2058.

[25] A. Bry and N. Roy, "Rapidly-exploring random belief trees for motion planning under uncertainty," in *2011 IEEE International Conference on Robotics and Automation*, 2011, pp. 723–730.

[26] S. Shi, X. Wang, and H. Li, "Pointrcnn: 3d object proposal generation and detection from point cloud," in *Proceedings of the IEEE Conference on Computer Vision and Pattern Recognition*, 2019, pp. 770–779.

[27] Y. Kuwata, J. Teo, G. Fiore, S. Karaman, E. Frazzoli, and J. P. How, "Real-time motion planning with applications to autonomous urban driving," *IEEE Transactions on control systems technology*, vol. 17, no. 5, pp. 1105–1118, 2009.

[28] A. Geiger, P. Lenz, and R. Urtasun, "Are we ready for autonomous driving? the kitti vision benchmark suite," in *2012 IEEE Conference on Computer Vision and Pattern Recognition*. IEEE, 2012, pp. 3354–3361.